%% file: main.tex
\documentclass[final]{cvpr}

\usepackage{times}
\usepackage{epsfig}
\usepackage{graphicx}
\usepackage{amsmath}
\usepackage{amssymb}

\usepackage{subfigure}
\usepackage{appendix}
\usepackage{xcolor}
\usepackage{color}
\usepackage{mathtools}
\usepackage{multirow}
\usepackage{pifont}
\usepackage[procnames]{listings}
\usepackage{wrapfig}
\usepackage[numbers,compress,sort]{natbib}
\usepackage{booktabs}
\usepackage{ulem}

\usepackage[pagebackref=true,breaklinks=true,colorlinks,bookmarks=false]{hyperref}

\def\confYear{CVPR 2021}

\begin{document}

\input{macro}
\title{On Feature Normalization and Data Augmentation}

\author{
Boyi Li$\phantom{}^{*12}$ \hspace{7pt} 
Felix Wu$\phantom{}^{*3}$ \hspace{7pt} 
Ser-Nam Lim$\phantom{}^{4}$ \hspace{7pt} 
Serge Belongie$\phantom{}^{12}$ \hspace{7pt} 
Kilian Q. Weinberger$\phantom{}^{13}$ \\
\vspace{2pt}
$\phantom{}^1$Cornell University \hspace{15pt} 
$\phantom{}^2$Cornell Tech \hspace{15pt} 
$\phantom{}^3$ASAPP \hspace{15pt} 
$\phantom{}^4$Facebook AI \\
{\tt\small \{bl728,sjb344,kilian\}@cornell.edu \hspace{7pt} 
fwu@asapp.com \hspace{7pt}
sernamlim@fb.com
}

}

\maketitle
\begin{abstract}
   \input{sections/abstract.tex}
\end{abstract}

\let\thefootnote\relax\footnotetext{$*$: Equal contribution.}
\input{sections/intro.tex}

\input{sections/related_work.tex}
\input{sections/method.tex}

\input{sections/experiments.tex}

\input{sections/discussion.tex}
\input{sections/conclusion.tex}

\input{sections/acknowledgement.tex}

{\small
\bibliographystyle{ieee_fullname}
\bibliography{reference}
}

\clearpage
\appendix
\input{sections/appendix.tex}

\end{document}

%% file: macro.tex
\newcommand{\methodnamefull}{Moment Exchange}
\newcommand{\methodname}{MoEx}

\definecolor{bleudefrance}{rgb}{0.0, 0.36, 1.0} 

\newcommand*{\shortautoref}[1]{%
  \begingroup
    \def\sectionautorefname{Sec.}%
    \def\subsectionautorefname{Subsec.}%
    \def\figureautorefname{Fig.}%
    \def\algorithmautorefname{Alg.}%
    \def\listingautorefname{List.}%
    \autoref{#1}%
  \endgroup
}

\renewcommand{\lstlistingname}{Algorithm}
\renewcommand{\lstlistlistingname}{List of \lstlistingname s}

\newcommand{\cmark}{\ding{51}}%
\newcommand{\xmark}{\ding{55}}%

%% file: sections/abstract.tex
The moments (a.k.a., mean and standard deviation) of latent features are often removed as noise when training image recognition models, to increase stability and reduce training time. However, in the field of image generation, the moments play a much more central role. Studies have shown that the moments extracted from instance normalization and positional normalization can roughly capture style and shape information of an image. Instead of being discarded, these moments are instrumental to the generation process. 
In this paper we propose Moment Exchange, an implicit data augmentation method that encourages the model to utilize the moment information also for recognition models.  
Specifically, we replace the moments of the learned features of one training image by those of another, and also interpolate the target labels---forcing the model to extract training signal from the moments in addition to the normalized features. 
As our approach is fast, operates entirely in feature space, and mixes different signals than prior  methods, one can effectively combine it with existing augmentation approaches. We demonstrate its efficacy across several recognition benchmark data sets 
where it improves the generalization capability of highly competitive baseline networks with remarkable consistency.

%% file: sections/intro.tex
\section{Introduction}
\label{sec:intro}
Image recognition and image generation are two corner stones of computer vision.
While both are burgeoning fields, specialized techniques from both sub-areas can sometimes form a dichotomy. Examples are 
mixup~\citep{zhang2017mixup} and squeeze-and-excitation~\citep{hu2018squeeze}
from the former, and adaptive instance normalization~\cite{huang2017arbitrary} from the latter, although exceptions exist.  Historically, the field of deep learning was widely popularized in discriminative image classification  with  AlexNet~\citep{krizhevsky2012imagenet}, and image generation through GANs~\citep{goodfellow2014gan} and VAEs~\citep{kingma2013vae}. 

\begin{figure}
    \centering
    \includegraphics[width=0.75\linewidth]{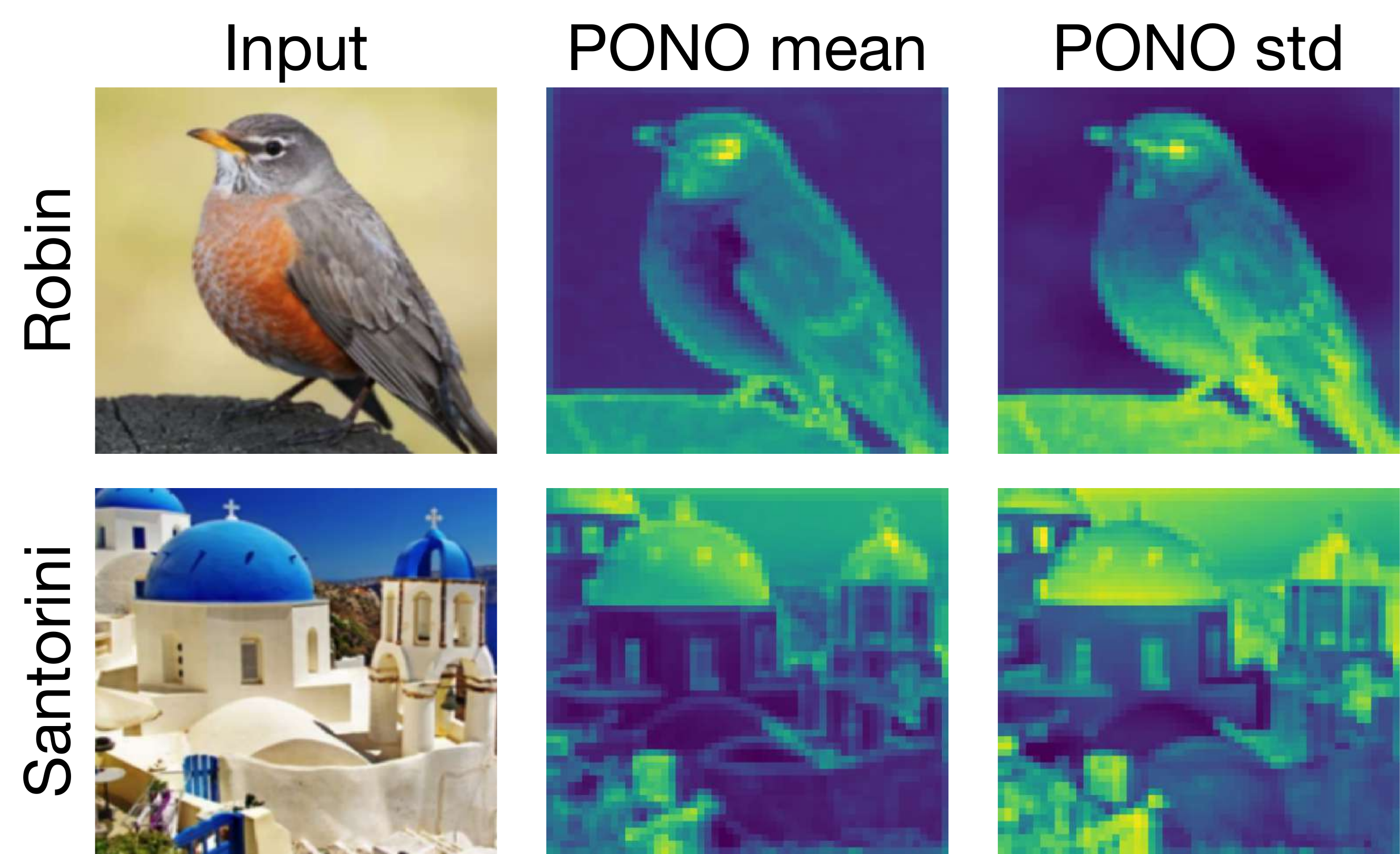}
    \caption{PONO mean and std captures structural information.}
    \label{fig:pono_show}
    \vspace{-0.2in}
\end{figure}

\begin{figure}
    \centering
    \includegraphics[width=1.0\linewidth]{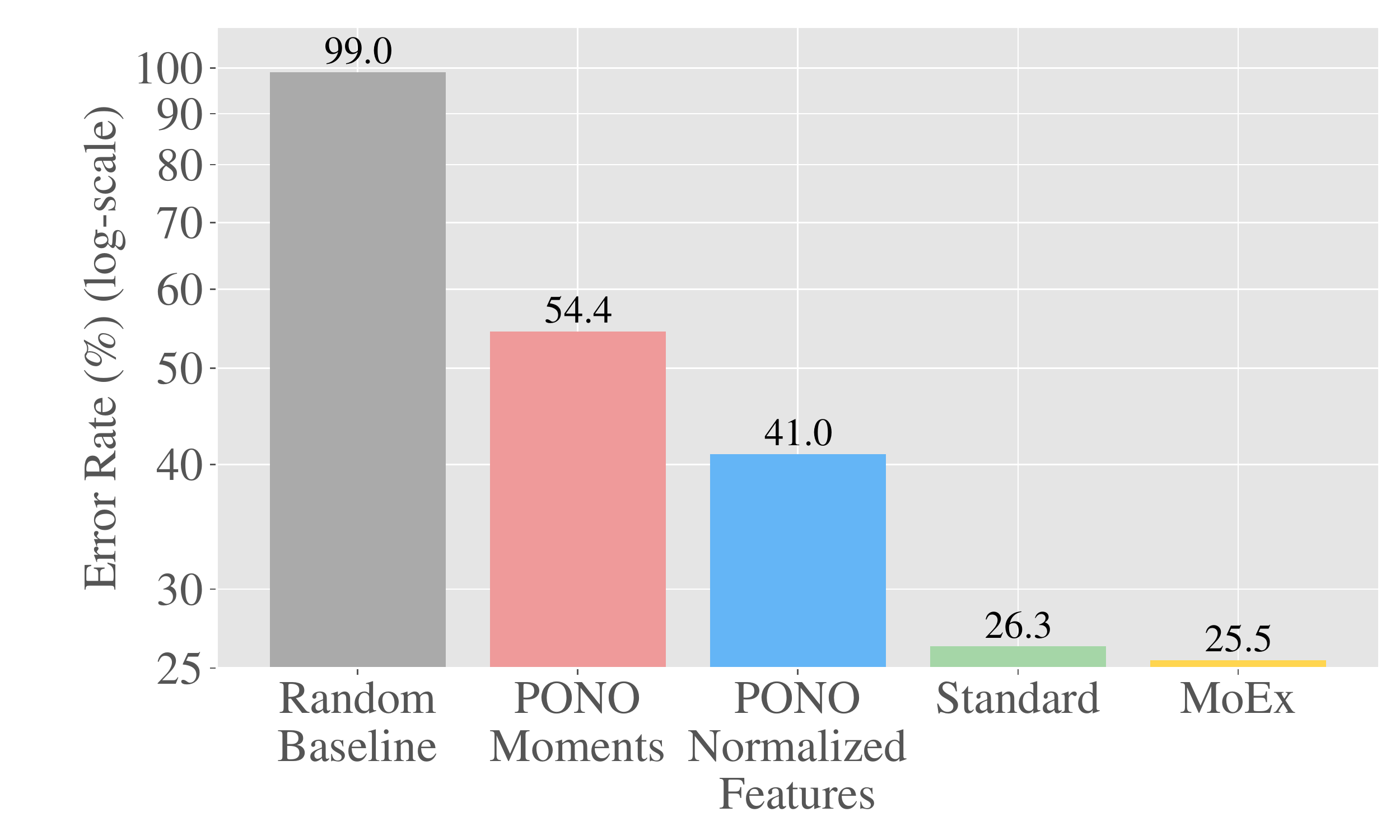}
    \caption{Error rates of ResNet-110 using different features on CIFAR-100. The numbers are averaged over three random runs. We compute the moments after the first convolutional layer, and either 
    adds  a  PONO  layer  right  after  the first Conv-BN-ReLU block or 
    computes the first two moments and concatenate them as a two-channel feature map. A ResNet which can only see the moments can still make nontrivial predictions (\textcolor[HTML]{EF9A9A}{Red} is much better than \textcolor[HTML]{AAAAAA}{gray}). Additionally, using only the normalized feature (i.e., removing the PONO moments) hurts the performance (\textcolor[HTML]{64B5F6}{Blue} is worse than \textcolor[HTML]{A5D6A7}{green}), which also shows that these moments contains important information. Finally, MoEx improves the performance by encouraging the model to use both sources of signal during training.
    }
    \label{fig:sep_err}
    \vspace{-0.2in}
\end{figure}

One particular aspect of this dichotomy is that in image recognition, popularized by batch normalization~\citep{ioffe2015batch}, the first and second moments (a.k.a., mean and standard deviation) in image recognition are computed across instances in a mini-batch and typically removed as noise~\citep{ioffe2015batch,wu2018group}. Studies have shown that this smoothes the optimization landscape~\cite{santurkar2018does} and enables larger learning rates~\cite{bjorck2018understanding}, which leads to faster convergence in practice. In contrast, for techniques like instance normalization~\citep{ulyanov2016instance} and positional normalization~\citep{li2019positional}, moments play a central role in the image generation process. For example, exchanging moments of latent features across samples has become a popular way to control for the style or shape of generated images~\citep{huang2017arbitrary,huang2018munit,karras2019style,park2019spade,li2019positional}.
Here, moments are viewed as features, not noise, with research showing that they encode the style of an image~\citep{huang2017arbitrary,karras2019style}, as well as the underlying structure~\citep{li2019positional}. To substantiate this point, we depict the first and second moments of the features extracted in the first layer of a ResNet~\citep{He2015} in \shortautoref{fig:pono_show}~\footnote{Example images are from Shutterstock.},  using the technique described in \cite{li2019positional}. The class label can still be inferred visually from both moments, which is a testament to the signal that remains in these statistics. To further substantiate our observation, we also show in \shortautoref{fig:sep_err} that simply using moments (from the first ResNet layer) for image classification already yields non-trivial performance (red bar) as compared to random guessing (gray bar). Similarly, removing the moments from positional normalization has a detrimental effect (blue bar vs. green bar).

As there is evidently important signal in the moments and in the normalized features, we would like to introduce a way to \textit{regulate} how much attention the deep net should pay to each source.  One approach to direct neural networks to a particular signal source is to introduce dedicated feature augmentation. For example, it has been shown that ConvNets trained on ImageNet~\citep{deng2009imagenet} are biased towards textures instead of shapes~\citep{geirhos2018imagenet}. To overcome this, \citet{geirhos2018imagenet} introduce a style transfer model to create a set of images with unreal textures. For example, they generate cats with elephant skins and bears with Coca-Cola bottle texture. An image classifier is trained to recognize the shape (cats or bears) instead of the textures (elephants or bottles).

\begin{figure*}[t]
    \centering
    \includegraphics[width=0.9\linewidth]{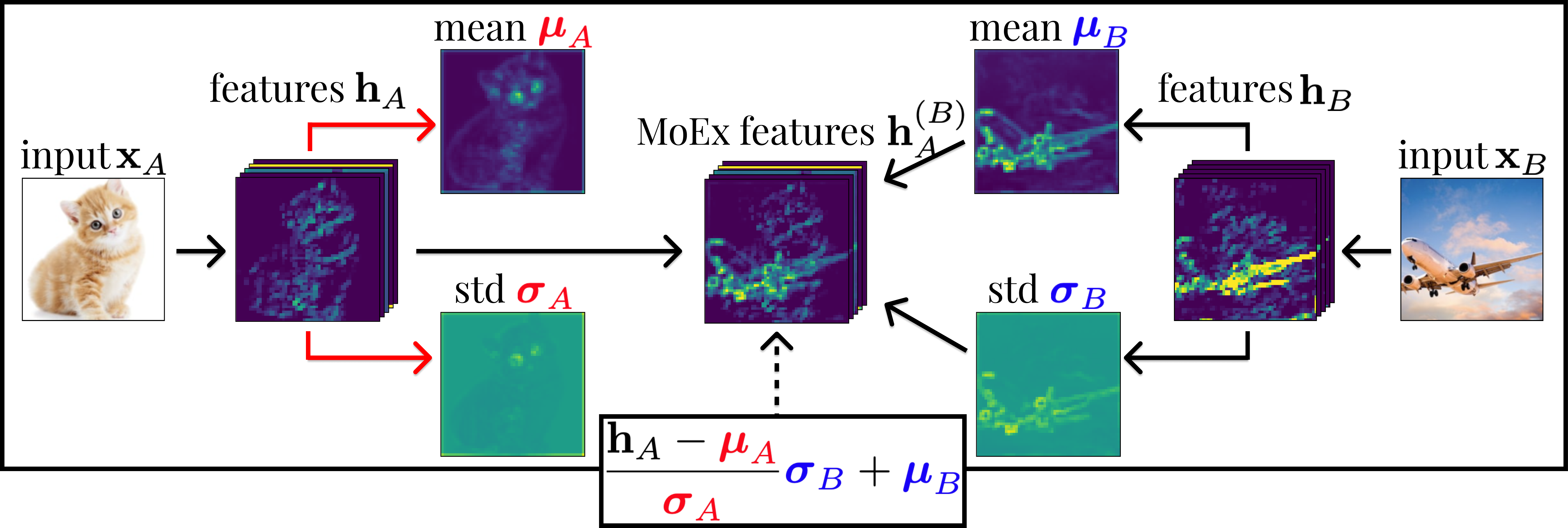}
    \caption{\methodname{} with PONO normalization. The features $\mathbf{h}_A$ of the cat image  are normalized and then infused with moments $\boldsymbol{\mu}_B,\boldsymbol{\sigma}_B$ from the plane image. See Appendix for more examples.}
    \label{fig:example_show1}
    \vspace{-0.2in}
\end{figure*}

In this paper we propose a novel data augmentation scheme that, to our knowledge, is the first method to systematically regulate how much attention a network pays to the signal in the feature moments. Concretely, we extract the mean and variance (across channels) after the first layer, but instead of simply removing them, we swap them between images.
See~\shortautoref{fig:example_show1} for a schematic illustration, where we extract and remove the feature moments of a cat image, and inject the moments of a plane.  
Knowing that the resulting features now contain information about both images, we make the network predict an interpolation of the two labels. 
In the process, we force the network to pay attention to two aspects of the data: the normalized feature (from the cat) and the moments (from the plane). 
By basing its prediction on two different signals we increase the robustness of the classification, as during testing both would point towards the same label. We call our method Moment Exchange or \methodname{} for short. 

Through exchanging the moments, we swap the shape (or style) information of two images, which can be viewed as an implicit version of the aforementioned method proposed by \citet{geirhos2018imagenet}.
However, \methodname{} does not require a pre-trained style transfer model to create a dataset explicitly.
In fact, \methodname{} is very effective for training with mini-batches and can be implemented in a few lines of code: During training we compute the feature mean and variance for each instance at a given layer (acros channels), permute them across the mini-batch, and re-inject them into the feature representation of other instances (while interpolating the labels). 

\methodname{} operates purely in feature space and can therefore easily be applied jointly with existing data augmentation methods that operate in the input space, such as cropping, flipping, rotating, but even label-perturbing approaches like Mixup~\citep{zhang2017mixup} or Cutmix~\citep{yun2019cutmix}. Importantly, because \methodname{} only alters the first and second moments of the pixel distributions, it has an orthogonal effect to existing data augmentation methods and its improvements can be ``stacked'' on top of their established gains in generalization. 
We conduct extensive experiments on eleven different tasks/datasets using more than ten varieties of models.
The results show that \methodname{} consistently leads to significant improvements across models and tasks, and it is particularly well suited to be combined with existing augmentation approaches. Further, our experiments show that \methodname{}  is not limited to  computer vision, but is also readily applicable and highly effective in applications within speech recognition and natural language processing---suggesting that \methodname{} reveals a fundamental insight about deep nets that crosses areas and data types. Our implementation is available at \href{https://github.com/Boyiliee/MoEx}{https://github.com/Boyiliee/MoEx}.

%% file: sections/related_work.tex
\section{Background and Related Work}
\label{sec:background}

Feature normalization has always been a prominent part of neural network training~\cite{lecun-98b,li1998sphering}. Initially, when networks had predominately one or two hidden layers, the practice of z-scoring the features was limited to the input itself. As networks became deeper, \citet{ioffe2015batch} extended the practice to the intermediate layers with the celebrated BatchNorm algorithm. As long as the mean and variance are computed across the entire input, or a randomly picked mini-batch (as it is the case for BatchNorm), the extracted moments reveal biases in the data set with no predictive information --- removing them causes no harm but can substantially improve optimization and generalization~\cite{lecun-98b,bjorck2018understanding,ross2013normalized}. 

In contrast, recently proposed normalization methods~\citep{ba2016layer,ulyanov2016instance,wu2018group,li2019positional} treat the features of each training instance as a distribution and normalize them for each sample individually. We refer to the extracted mean and variance as \textit{intra-instance} moments. 
We argue that intra-instance moments are attributes of a data instance that describe the distribution of its features and should not be discarded. 
Recent works~\citep{huang2017arbitrary,li2019positional} have shown that such attributes can be useful in several generative models. 
Realizing that these moments capture interesting information about data instances, we propose to use them for data augmentation.

Data augmentation has a similarly long and rich history in machine learning. Initial approaches discovered the concept of \textit{label-preserving} transformations~\citep{simard1993efficient,scholkopf1996incorporating} to mimic larger training data sets to suppress overfitting effects and improve  generalization. 
For instance, \citet{simard2003best} randomly translate or rotate images assuming that the labels of the images would not change under such small perturbations. 
Many subsequent papers proposed alternative flavors of this augmentation approach based on similar insights~\citep{devries2017cutout,dualcutout,cubuk2019autoaugment,zhong2020random,karras2019style,cubuk2019randaugment,xie2019adversarial,singh-iccv2017,li2020shapetexture}.
Beyond vision tasks, back-translation~\citep{sennrich2015improving,yu2018qanet,edunov2018understanding,caswell2019tagged} and word dropout~\citep{iyyer2015deep} are commonly used to augment text data. Besides augmenting inputs, \citet{maaten2013learning,ghiasi2018dropblock,wang2019implicit} adjust either the features or loss function as implicit data augmentation methods.
In addition to label-preserving transformations, there is an increasing trend to use \textit{label-perturbing} data augmentation methods.
\citet{zhang2017mixup} arguably pioneered the field with Mixup, which interpolates two training inputs in feature and label space simultaneously. Cutmix~\citep{yun2019cutmix}, instead, is designed especially for image inputs. It randomly crops a rectangular region of an image and pastes it into another image, mixing the labels proportional to the number of pixels contributed by each input image to the final composition. 

%% file: sections/method.tex
\section{\methodnamefull{}}
\label{sec:method}

In this section we introduce \methodnamefull{} (\methodname{}), which blends feature normalization with data augmentation. Similar to Mixup and Cutmix, it fuses features and labels across two training samples, however it is unique in its asymmetry, as it mixes two very different components: The normalized features of one instance are combined with the feature moments of another. This asymmetric composition in feature space  allows us to capture and smooth out different directions of the decision boundary, not previously covered by existing augmentation approaches. We also show that \methodname{} can be implemented very efficiently in a few lines of code, and should be regarded as an effective default companion to existing data augmentation methods. 

\textbf{Setup.}
Deep neural networks are composed of layers of transformations including convolution, pooling, transformers~\cite{vaswani2017attention}, fully connected layers, and non-linear activation layers. Consider a batch of input instances $\mathbf{x}$, these transformations are applied sequentially to generate a series of hidden features $\mathbf{h}^1,  ..., \mathbf{h}^L$ before passing the final feature $\mathbf{h}^L$ to a linear classifier. 
For each instance, any feature presentation $\mathbf{h}^\ell$ is a 3D tensor indexed by channel (C), height (H), and width (W).

\textbf{Normalization.}
We assume the network is using an invertible \emph{intra-instance} normalization. 
We denote this function by $F$, which takes the features $\mathbf{h}_i^\ell$ of the $i$-th input $\mathbf{x}_i$ at layer $\ell$ and produces three outputs: the normalized features $\hat{\mathbf{h}}_i$, the first moment $\boldsymbol{\mu}_i$, and the second moment $\boldsymbol{\sigma}_i$:
\[
(\hat{\mathbf{h}}_i^\ell, \boldsymbol{\mu}_i^\ell, \boldsymbol{\sigma}_i^\ell) = F(\mathbf{h}_i^\ell),\ \  \mathbf{h}_i^\ell = F^{-1}(\hat{\mathbf{h}}^\ell_i, \boldsymbol{\mu}^\ell_i, \boldsymbol{\sigma}^\ell_i).
\]
The inverse function $F^{-1}$ reverses the normalization process. 
As an example, PONO~\citep{li2019positional} computes the first and second moments across channels from the feature representation at a given layer
\begin{align*}
\boldsymbol{\mu}_{b,h,w}^\ell 
&= \frac{1}{C} \sum_{c} \mathbf{h}_{b,c,h,w}^\ell, \\  \boldsymbol{\sigma}_{b,h,w}^\ell
&= \sqrt{ \frac{1}{C}\sum_{c} \left(\mathbf{h}_{b, c, h, w}^\ell - \boldsymbol{\mu}_{b,h,w}^\ell\right)^2 + \epsilon}.
\end{align*}
The normalized features have zero-mean and standard deviation 1 along the channel dimension. Note that after using MoEx with an \textit{intra-instance} normalization to exchange features,  we can still apply  an \textit{inter-instance} normalization (like BatchNorm) on these exchanged or mixed features, with their well-known beneficial impact on convergence. As the norms compute statistics across different dimensions their interference is insignificant.

\textbf{\methodnamefull{}.} 
The procedure described in the following functions identically for each layer it is applied to and we therefore drop the $\ell$ superscript for notational simplicity. Further, for now, we only consider two randomly chosen samples $\mathbf{x}_A$ and $\mathbf{x}_B$ (see \shortautoref{fig:example_show1} for a schematic illustration). The intra-instance normalization decomposes the features of input $\mathbf{x}_A$ at layer $\ell$ into three parts, $\hat{\mathbf{h}}_A, \boldsymbol{\mu}_A, \boldsymbol{\sigma}_A$. Traditionally, batch-normalization~\cite{ioffe2015batch} discards the two moments and only proceeds with the normalized features $\hat{\mathbf{h}}_A$. If the moments are computed across instances (e.g. over the mini-batch) this makes sense, as they capture biases that are independent of the label. However, in our case we focus on intra-instance normalization (See \shortautoref{fig:pono_show}), and therefore both moments are computed only from $\mathbf{x}_A$ and are thus likely to contain label-relevant signal. This is clearly visible in the \textit{cat} and \textit{plane} examples in  \shortautoref{fig:example_show1}. All four moments ($\boldsymbol{\mu}_A$,$\boldsymbol{\sigma}_A$,$\boldsymbol{\mu}_B$,$\boldsymbol{\sigma}_B$), capture the underlying structure of the samples,  revealing their respective class labels.

We consider the normalized features and the moments as distinct views of the same instance. It generally helps robustness if a machine learning algorithm leverages multiple sources of signal, as it becomes more resilient in case one of them is under-expressed in a test example. For instance, the first moment conveys primarily structural information and only little color information, which, in the case of cat images can help overcome overfitting towards fur color biases in the training data set. 

In order to encourage the network to utilize the moments, we use the two images and combine them by injecting the moments of image $\mathbf{x}_B$ into the feature representation of image $\mathbf{x}_A$:
\(
    \mathbf{h}_A^{(B)}= F^{-1}({\hat{\mathbf{h}}}_A,\boldsymbol{\mu}_B,\boldsymbol{\sigma}_B).
\)
In the case of PONO, the transformation becomes
\(
    \mathbf{h}_A^{(B)} = 
    \boldsymbol{\sigma}_B \frac{\mathbf{h}_A - \boldsymbol{\mu}_A}{\boldsymbol{\sigma}_A}  + \boldsymbol{\mu}_B.
\)
We now proceed with these features $\mathbf{h}^{(B)}_A$, which contain the moments of image B (plane) hidden inside the features of image A (cat).
In order to encourage the neural network to pay attention to the injected features of B we modify the loss function to predict the class label $y_A$ and also $y_B$, up to some mixing constant  $\lambda \in [0, 1]$.
The loss becomes a straight-forward combination
\[
\lambda \cdot \ell(\mathbf{h}_{A}^{(B)}, y_{A}) + (1 - \lambda) \cdot \ell(\mathbf{h}_{A}^{(B)}, y_{B}).\label{eq:smoothbaby}
\]

\textbf{Implementation.}
In practice one needs to apply \methodname{} only on a single layer in the neural network, as the fused signal is propagated until the end. With PONO as the normalization method, we observe that the first layer ($\ell=1$) usually leads to the best result. In contrast, we find that \methodname{} is more suited for later layers when using IN~\citep{ulyanov2016instance}, GN~\citep{wu2018group}, or LN~\citep{ba2016layer} for moment extraction. 
Please see \shortautoref{sec:discussion} for a detailed ablation study. 
The inherent randomness of mini-batches allows us to implement \methodname{} very efficiently. For each input instance in the mini-batch $\mathbf{x}_i$ we compute the normalized features and moments  $\hat{\mathbf{h}}_i,\boldsymbol{\mu}_i,\boldsymbol{\sigma}_i$. Subsequently we sample a random permutation $\pi$ and apply \methodname{} with a random pair within the mini-batch
\(
    \mathbf{h}_i^{\left(\pi(i)\right)}\leftarrow 
    F^{-1}({\hat{\mathbf{h}}}_i,\boldsymbol{\mu}_{\pi(i)},\boldsymbol{\sigma}_{\pi(i)}).
\)
See Algorithm 1 in the Appendix for an example implementation in PyTorch~\citep{paszke2017automatic}. Note that all computations are extremely fast and only introduce negligible overhead during training.

\textbf{Hyper-parameters.}
To control the intensity of our data augmentation, we perform \methodname{} during training with some probability $p$. In this way, the model can still see the original features with probability $1 - p$. In practice we found that $p = 0.5$ works well on most datasets except that we set $p = 1$ for ImageNet where we need stronger data augmentation.
The interpolation weight $\lambda$ is another hyper-parameter to be tuned. Empirically, we find that 0.9 works well across data sets. The reason can be that the moments contain less information than the normalized features.
Please see Appendix for a detailed ablation study.

\textbf{Properties.} \methodname{} is performed entirely at the feature level inside the neural network and can be readily combined with other augmentation methods that operate on the raw input (pixels or words). For instance,  Cutmix~\cite{yun2019cutmix} typically works best when applied on the input pixels directly. We find that the improvements of \methodname{} are complimentary to such prior work and recommend to use \methodname{} in combination with established data augmentation methods. 

\vspace{-0.1in}

%% file: sections/experiments.tex
\section{Experiments}
\label{sec:exp}

We evaluate the efficacy of \methodname{} thoroughly across several tasks and data modalities. Our implementation will be released as open source upon publication.

\subsection{Image Classification on CIFAR}

\begin{table}[t]
    \centering
    \input{tables/moex_cifar100_result1}
    \vspace{-0.1in}
\end{table}

\begin{table}[t]
    \centering
    \input{tables/moex_cifar100_result2}
    \vspace{-0.2in}
\end{table}

\label{subsec:cifar}
CIFAR-10 and CIFAR-100~\cite{krizhevsky2009learning} are benchmark datasets containing 50K training and 10K test colored images at 32x32 resolution. We evaluate our method using various model architectures~\cite{He2015,huang2017densely,xie2017aggregated,zagoruyko2016wide,han2017deep} on CIFAR-10 and CIFAR-100. We follow the conventional setting\footnote{\href{https://github.com/bearpaw/pytorch-classification}{https://github.com/bearpaw/pytorch-classification}} with random translation as the default data augmentation and apply \methodname{} to the features after the first layer. 
Furthermore, to justify the compatibility of \methodname{} with other regularization methods, we follow the official setup\footnote{\href{https://github.com/clovaai/CutMix-PyTorch}{https://github.com/clovaai/CutMix-PyTorch}} of \citep{yun2019cutmix} and apply \methodname{} jointly with several regularization methods to PyramidNet-200~\cite{han2017deep} on CIFAR-100.

\autoref{tab:moex_cifar100_result1} displays the classification results on CIFAR-10 and CIFAR-100 with and without \methodname{}. We report mean and standard error over three runs~\cite{gurland1971simple}. \methodname{} consistently enhances the performance of all the baseline models. 
\autoref{tab:moex_cifar100_result2} demonstrates the CIFAR-100 classification results on the basis of PyramidNet-200. 
Compared to other augmentation methods, PyramidNet trained with \methodname{} obtains the lowest error rates in all-but one settings. However, significant additional improvements are achieved when \methodname{}  is combined with existing methods  --- setting a new  state-of-the-art for this particular benchmark task when combined with the two best performing alternatives, CutMix and ShakeDrop.

\subsection{Image Classification on ImageNet}

\begin{table}[t]
    \centering
    \input{tables/moex_imagenet_baseline_result}
\end{table}

\begin{table}[t]
    \centering
    \input{tables/imagenet_da_result}
    \vspace{-0.2in}
\end{table}

\label{subsec:imagenet}
We evaluate on ImageNet~\cite{deng2009imagenet} (ILSVRC 2012 version), which consists of 1.3M training images and 50K validation images of various resolutions. For faster convergence, we use NVIDIA's mixed-precision training code base\footnote{\href{https://github.com/NVIDIA/apex/tree/master/examples/imagenet}{https://github.com/NVIDIA/apex/tree/master/examples/imagenet}} with batch size 1024, default learning rate $0.1 \times \mbox{batch\_size} / 256$, cosine annealing learning rate scheduler~\cite{loshchilov2016sgdr} with linear warmup~\cite{goyal2017accurate} for the first 5 epochs. 
As the model might require more training updates to converge with data augmentation, we apply \methodname{} to ResNet-50, ResNeXt-50 (32$\times$4d), DenseNet-265 and train them for 90 and 300 epochs. For a fair comparison, we also report Cutmix~\cite{yun2019cutmix} under the same setting.
Since the test server of ImageNet is no longer available to the public, we follow the common practice~\cite{xie2017aggregated,huang2017densely,zhang2017mixup,yun2019cutmix,ghiasi2018dropblock} reporting the validation scores of the baselines as well as our method.

\autoref{tab:moex_imagenet_baseline_result} shows the test error  rates on the ImageNet data set. 
\methodname{} is able to improve the classification performance throughout, regardless of model architecture. Similar to the previous CIFAR experiments, we observe in  \autoref{tab:imagenet_da_result} that \methodname{} is  highly competitive when compared to existing regularization methods and truly shines when it  is combined with them. When applied jointly with CutMix (the  strongest alternative), we obtain our lowest Top-1 and Top-5 error of 20.9/5.7  respectively. Beyond, we apply \methodname{} to   EfficientNet-B0~\cite{tan2019efficientnet} and follow the official training scheme, we find \methodname{} is able to help reduce the error rate of baseline from 22.9 to 22.3.

\subsection{Fintuneing Imagenet pretrained models on Pascal VOC for Object Detection}
\label{object_dectection}
To demonstrate that \methodname{} encourages models to learn better image representations, we apply models pre-trained on ImageNet with \methodname{} to downstream tasks including object detection on Pascal VOC 2007 dataset.
We use the Faster R-CNN~\citep{ren2015faster} with C4 or FPN~\citep{lin2017feature} backbones implemented in Detectron2~\citep{wu2019detectron2} and following their default training configurations.
We consider three ImageNet pretrained models: the ResNet-50 provided by \citet{He2015}, our ResNet-50 baseline trained for 300 epochs, our ResNet-50 trained with CutMix~\citep{yun2019cutmix}, and our ResNet-50 trained with \methodname{}.
A Faster R-CNN is initialized with these pretrained weights and finetuned on Pascal VOC 2007 + 2012 training data, tested on Pascal VOC 2007 test set, and evaluated with the PASCAL VOC style metric: average precision at IoU 50\% which we call AP\textsubscript{VOC} (or AP50 in detectron2). We also report MS COCO~\citep{lin2014microsoft} style average precision metric AP\textsubscript{COCO} which is recently considered as a better choice.
Notably, \methodname{} is not applied during finetuning.

\autoref{tab:voc} shows the average precision of different initializations. We discover that \methodname{} provides a better initialization than the baseline ResNet-50 and is competitive against CutMix\citep{yun2019cutmix} for the downstream cases and leads slightly better performance regardless of backbone architectures. 

\begin{table}[h]
    \centering
    \input{tables/voc}
    \vspace{-0.15in}
\end{table}

\subsection{3D model classification on ModelNet}
\label{subsec:modelnet}
We conduct experiments on Princeton ModelNet10 and ModelNet40 datasets~\cite{wu20153d} for 3D model classification. This task aims to classify 3D models encoded as 3D point clouds into 10 or 40 categories. As a proof of concept, we use PointNet++ (SSG)~\cite{qi2017pointnet++} implemented efficiently in PyTorch Geometric\footnote{\href{https://github.com/rusty1s/pytorch_geometric/blob/master/examples/pointnet2_classification.py}{https://github.com/rusty1s/pytorch\_geometric}}~\cite{Fey/Lenssen/2019} as the baseline. It does not use surface normal as additional inputs.
We apply \methodname{} to the features after the first set abstraction layer in PointNet++. Following their default setting, all models are trained with ADAM~\cite{kingma2014adam} at batch size 32 for 200 epochs. The learning rate is set to 0.001. We tune the hyper-parameters of \methodname{} on ModelNet-10 and apply the same hyper-parameters to ModelNet-40.  We choose $p = 0.5$, $\lambda = 0.9$, and InstanceNorm\footnote{We do hyper-parameter search from $p \in \{0.5, 1\}$, $\lambda \in \{0.5, 0.9\}$ and whether to use PONO or InstanceNorm. } for this task, which leads to slightly better results.

\autoref{tab:pointnet} summarizes the results out of three runs, showing mean error rates with standard errors.  \methodname{} reduces the classification errors from 6.0\% to 5.3\% and 9.2\% to 8.8\% on ModelNet10 and ModelNet40, respectively.

\begin{table}[h]
    \centering
    \input{tables/pointnet_result}
    \vspace{-0.2in}
\end{table}

%% file: tables/moex_cifar100_result1.tex
\footnotesize
\begin{tabular}{c|c|c|c}
\toprule
Model & \#param. & CIFAR10 & CIFAR100 \\  
\midrule\midrule
ResNet-110 (3-stage) & 1.7M &  6.82$\pm$0.23& 26.28$\pm$0.10 \\
+\methodname{} & 1.7M & \textbf{6.03$\pm$0.24}& \textbf{25.47$\pm$0.09} \\

\midrule
DenseNet-BC-100 (k=12) & 0.8M & 4.67$\pm$0.10
& 22.61$\pm0.17$   \\
+\methodname{} &0.8M&\textbf{4.58$\pm$0.03} &  \textbf{21.38$\pm$0.18}\\
\midrule
ResNeXt-29 (8$\times$64d) & 34.4M &4.00$\pm$0.04 &18.54$\pm$0.27  \\
+\methodname{} &34.4M&\textbf{3.64$\pm$0.07}&\textbf{17.08$\pm$0.12}\\
\midrule
WRN-28-10 & 36.5M &  3.85$\pm$0.06& 18.67$\pm0.07$ \\
+\methodname{} & 36.5M & \textbf{3.31$\pm$0.03} &  \textbf{17.69$\pm$0.10} \\
\midrule
DenseNet-BC-190 (k=40) & 25.6M &3.31$\pm$0.04 &17.10$\pm$0.02  \\
+\methodname{} &25.6M&\textbf{2.87$\pm$0.03}& \textbf{16.09$\pm$0.14}\\
\midrule
PyramidNet-200 ($\tilde{\alpha}=240$) & 26.8M &3.65$\pm$0.10  & 16.51$\pm0.05$ \\
+\methodname{} & 26.8M & \textbf{3.44$\pm$0.03} &  \textbf{15.50$\pm$0.27} \\
\bottomrule
\end{tabular}
\caption{
Classification results (Err (\%)) on CIFAR-10, CIFAR-100 in comparison with various competitive baseline models. See text for details.
}
\label{tab:moex_cifar100_result1}

%% file: tables/moex_cifar100_result2.tex
\begin{tabular}{l|c}
\toprule
PyramidNet-200 ($\tilde{\alpha}=240$) &Top-1 / Top-5\\
(\# params: 26.8 M) & Error (\%)\\
\midrule
\midrule
    Baseline &16.45 / 3.69 \\
    Manifold Mixup~\cite{zhang2017mixup}&16.14 / 4.07 \\
    StochDepth~\cite{huang2016deep}& 15.86 / 3.33\\
    DropBlock~\cite{ghiasi2018dropblock}&15.73 / 3.26 \\
    Mixup~\cite{zhang2017mixup}&15.63 / 3.99 \\
    ShakeDrop~\cite{yamada2018shakedrop} & 15.08 /\textbf{ 2.72}\\
    \methodname{}&\textbf{15.02} / 2.96\\
\midrule
    Cutout~\cite{devries2017cutout}&16.53 / 3.65  \\
    Cutout + \methodname{}& \textbf{15.11} / \textbf{3.23} \\
    \midrule
    CutMix~\cite{yun2019cutmix}& 14.47 / 2.97\\
    CutMix + \methodname{}& \textbf{13.95} / \textbf{2.95} \\
    \midrule
    CutMix + ShakeDrop~\cite{yamada2018shakedrop} & 13.81 / 2.29\\
    CutMix + ShakeDrop + \methodname{}&    \textbf{13.47} / \textbf{2.15} \\
\bottomrule
\end{tabular}
\caption{Combining \methodname{} with other regularization methods on CIFAR-100 following the setting of~\cite{yun2019cutmix}. The best numbers in each group are \textbf{bold}.}
\label{tab:moex_cifar100_result2}

%% file: tables/moex_imagenet_baseline_result.tex
\begin{tabular}{c|c|c|c}
\toprule
& \multicolumn{1}{c}{\# of} & \multicolumn{2}{|c}{Test  Error (\%)} \\
\cline{3-4}
Model & epochs & Baseline &+\methodname{}  \\  
\midrule\midrule
ResNet-50   & 90 &23.6 &\textbf{23.1}  \\
ResNeXt-50 (32$\times$4d)& 90 & 22.2 &\textbf{21.4}   \\
DenseNet-265 &90&21.9 & \textbf{21.6}    \\
\midrule
ResNet-50 &  300& 23.1&\textbf{21.9} \\
ResNeXt-50 (32$\times$4d) &300 & 22.5 &\textbf{22.0}  \\
DenseNet-265 &300& 21.5&\textbf{20.9}  \\
\bottomrule
\end{tabular}
\caption{Classification results (Test Err (\%)) on ImageNet in comparison with various models.
Note: The ResNeXt-50 (32$\times$4d) models trained for 300 epochs overfit. They have higher training accuracy but lower test accuracy than the 90-epoch ones.}
\label{tab:moex_imagenet_baseline_result}

%% file: tables/imagenet_da_result.tex
\begin{tabular}{l|c|c}
    \toprule
    ResNet50  & \# of & Top-1 / Top-5\\
   (\# params:  25.6M) & epochs & Error (\%)\\
    \midrule
    \midrule
    ISDA~\cite{wang2019implicit} & 90 & 23.3 / 6.8 \\
    Shape-ResNet~\cite{geirhos2018imagenet} & 105 & 23.3 / 6.7 \\
    Mixup~\cite{zhang2017mixup} & 200 & 22.1 / 6.1  \\
    AutoAugment~\cite{cubuk2019autoaugment} & 270 & 22.4 / 6.2 \\
    Fast AutoAugment~\cite{lim2019fast} & 270 & 22.4 / 6.3 \\
    DropBlock~\cite{ghiasi2018dropblock} & 270 & 21.9 / 6.0 \\
    Cutout~\cite{devries2017cutout}& 300 & 22.9 / 6.7  \\ 
    Manifold Mixup~\cite{zhang2017mixup} & 300 & 22.5 / 6.2 \\
    Stochastic Depth~\cite{huang2016deep} & 300 & 22.5 / 6.3 \\ 
    CutMix~\cite{yun2019cutmix} & 300 & 21.4 / 5.9 \\
    \midrule
    Baseline & 300 & 23.1 / 6.6 \\  
    \methodname{} & 300 & 21.9 / 6.1 \\
    CutMix & 300 & 21.3 / \textbf{5.7} \\
    CutMix + \methodname{} & 300 & \textbf{20.9} / \textbf{5.7} \\
    \bottomrule
\end{tabular}
\caption{Comparison of regularization and augmentation methods on ImageNet. Stochastic Depth and Cutout results are from \cite{yun2019cutmix}.}
\label{tab:imagenet_da_result}

%% file: tables/voc.tex
\small
\begin{tabular}{c|l|c|c}
    \toprule
    Backbone & Initialization & AP\textsubscript{VOC} & AP\textsubscript{COCO} \\
    \midrule
    \multirow{3}{*}{C4}
    & ResNet-50 (default)  & 80.3  & 51.8 \\
    & ResNet-50 (300 epochs) & 81.2 & 53.5\\
    & ResNet-50 + CutMix & \textbf{82.1} & 54.3 \\
    & ResNet-50 + \methodname{} & 81.6 & \textbf{54.6} \\
    \midrule
    \multirow{3}{*}{FPN}
    & ResNet-50 (default) & 81.8 & 53.8 \\
    & ResNet-50 (300 epochs) & 82.0 & 54.2 \\
    & ResNet-50 + CutMix & 82.1 & \textbf{54.3}  \\
    & ResNet-50 + \methodname{} & \textbf{82.3} & \textbf{54.3} \\
    \bottomrule
\end{tabular}
\caption{Object detection on PASCAL VOC 2007 test set using Faster R-CNN whose backbone is initialized with different pretrained weights. We use either the original C4 or feature pyramid network~\citep{lin2017feature} backbone.}
\label{tab:voc}

%% file: tables/pointnet_result.tex
\begin{tabular}{l|c|c}
    \toprule
    Model & ModelNet10 & ModelNet40 \\
    \midrule
    \midrule
    PointNet++ & 6.02$\pm$0.10 & 9.16$\pm$0.16\\
    + \methodname{} & \textbf{5.25$\pm$0.18 }& \textbf{8.78$\pm$0.28}\\
    \bottomrule
\end{tabular}
\caption{Classification errors (\%) on ModelNet10 and ModelNet40. The mean and standard error out of 3 runs are reported.}
\label{tab:pointnet}

%% file: sections/discussion.tex
\section{Ablation Study}
\label{sec:discussion}

\begin{table}[t]
    \centering
    \input{tables/label_smoothing_ablation}
    \vspace{-0.2in}
\end{table}
\subsection{MoEx Design Choices}
In the previous section we have established that \methodname{} yields significant improvements across many tasks and model architectures. In this section we shed light onto which design choices crucially  contribute to these  improvements. \autoref{tab:label_smoothing_ablation} shows results  on  CIFAR-100 with a ResNet-110 architecture, averaged over 3 runs. 
The column titled MoEx indicates if we performed moment exchange or not. 
\vspace{-0.15in}
\paragraph{Label smoothing.}
First, we investigate if the positive effect of \methodname{} can be attributed to label smoothing~\cite{szegedy2016rethinking}. In label smoothing, one changes the loss of a sample $\mathbf{x}$ with label $y$ to
\(
\lambda\ell(\mathbf{x}, y) + \frac{1}{C-1}\sum_{y'\neq y} (1-\lambda) \ell(\mathbf{x}, y'),
\)
where $C$ denotes the total number of classes. Essentially the neural network is not trained to predict one class with 100\% certainty, but instead only up to a confidence $\lambda$. 

Further, we evaluate \textbf{Label Interpolation only.} Here, we evaluate \methodname{} with label interpolation - but without any feature augmentation,  essentially  investigating  the effect of label interpolation alone. Both variations yield some improvements over the baseline, but are significantly worse than \methodname{}. 
\vspace{-0.15in}
\paragraph{Interpolated targets.}
The last three rows of \autoref{tab:label_smoothing_ablation} demonstrate the necessity of utilizing the moments for prediction. We investigate two variants: $\lambda=1$, which corresponds  to no label interpolation; \methodname{} with label smoothing (essentially assigning a small loss to all  labels except $y_A$). 
The last row corresponds to our  proposed method, \methodname{} ($\lambda=0.9$). 

Two general observations can be made: 1) interpolating the labels is crucial for \methodname{} to be beneficial --- the approach leads to absolutely no improvement when we set $\lambda=1$. 2) 
 it is also important to perform moment exchange, without it \methodname{} reduces to a version of label smoothing, which yields significantly smaller benefits.  

\vspace{-0.15in}
\paragraph{Choices of normalization methods.}
We study how \methodname{} performs when using moments from LayerNorm (LN)~\cite{ba2016layer}, InstanceNorm (IN)~\cite{ulyanov2016instance}, PONO~\cite{li2019positional}, GroupNorm (GN)~\cite{wu2018group}, and local response normalization (LRN)~\cite{krizhevsky2012imagenet} perform.
For LRN, we use a recent variant~\cite{karras2018progressive} which uses the unnormalized 2nd moment at each position.
We conduct experiments on CIFAR-100 with ResNet110. For each normalization, we do a hyper-parameter sweep to find the best setup\footnote{We select the best result from experiments with $\lambda \in \{0.6, 0.7, 0.8, 0.9\} $ and $p \in \{0.25, 0.5, 0.75, 1.0\}$. We choose the best layer among the 1st layer, 1st stage, 2nd stage, and 3rd stage. For each setting, we obtain the mean and standard error out of 3 runs with different random seeds.}. 
\autoref{tab:moex_norm_ablation} shows classification results of \methodname{} with various feature normalization methods on CIFAR-100 averaged over 3 runs (with corresponding standard errors). We observe that \methodname{} generally works with all normalization approaches, however PONO has a slight but significant edge, which we attribute to the fact that it catches the structural information of the feature most effectively.
\autoref{tab:ablation_norm_layer} shows that different normalization methods work the best at different layers. With PONO or GN, we apply \methodname{} in the first layer (right before the first stage), whereas the LN moments work best when exchanged before the third stage of a 3-stage ResNet-110; IN is better to be applied right before the second stage. We hypothesize the reason is that PONO moments captures local information while LN and IN compute global features which are better encoded at later stages of a ResNet. For image classification, using PONO seems generally best. While MoEx with other normalization methods in different stages could also obtain competitive results such as LN before Stage 3 in \autoref{tab:moex_norm_ablation}. We assume it is because in the early layers it is important to exchange the whole mode (and PONO has a significant advantage), whereas for the last stage the scale already contains a lot of information (LN performs best here), which is worth being studied in other architecture such as Transformer~\cite{vaswani2017attention}, etc. Beyond, for some other tasks we observe that using moments from IN can be more favorable (See \shortautoref{subsec:modelnet}). 

\begin{table}[h]
    \centering
    \input{tables/moex_norm_ablation}
    \vspace{-0.25in}
\end{table}

\begin{table}[h]
    \centering
    \input{tables/ablation_norm_layer}

    \vspace{-0.2in}
\end{table}

\subsection{MoEx Hyper-parameters}
\label{subsec:ablation_hyper}
In \methodname{}, $\lambda$ and $1 - \lambda$ serve as the target interpolation weights of labels $y_A$, $y_B$, respectively. To  explore the relationship between $\lambda$ and model performance, we train a ResNet-50 on ImageNet with $\lambda \in \{0.3, 0.5, 0.7, 0.9\}$ with on PONO. The results are summarized in  \autoref{tab:ablation_choices}. 
We observe  that generally higher $\lambda$ leads to lower error, probably because more information is captured in the normalized features than in the moments.  
After all, moments only capture general statistics, whereas the features have many channels and can capture texture information in great detail. 
We also investigate various values of the exchange probability $p$ (for fixed $\lambda=0.9$), but on the ImageNet data $p=1$ (i.e. apply \methodname{} on every image) tends to  perform best. 

\begin{table}[h]
    \centering
    \input{tables/ablation_choices}
    \vspace{-0.2in}
\end{table}

\subsection{Model Analysis}
To estimate the robustness of the models trained with \methodname{}, we follow  the procedure proposed by~\cite{hendrycks2019natural} and evaluate our model on their ImageNet-A data set, which  contains 7500 natural images (not originally part of  ImageNet) that are misclassified by a publicly released ResNet-50 in torchvision\footnote{\href{https://download.pytorch.org/models/resnet50-19c8e357.pth}{https://download.pytorch.org/models/resnet50-19c8e357.pth}}. 
We compare our models with various publicly released pretrained models including Cutout~\cite{zhang2017mixup}, Mixup~\cite{zhang2017mixup}, CutMix~\cite{yun2019cutmix}, Shape-ResNet~\cite{geirhos2018imagenet}, and recently proposed AugMix~\cite{hendrycks2020augmix}.
We report all 5 metrics implemented in the official evaluation code\footnote{\href{https://github.com/hendrycks/natural-adv-examples}{https://github.com/hendrycks/natural-adv-examples}}: model accuracy (Acc), root mean square calibration rrror (RMS), mean absolute distance calibration error (MAD), the area under the response rate accuracy curve (AURRA) and soft F1~\cite{sokolova2006beyond, hendrycks2019natural}. 
\autoref{tab:imagenet_a} summarizes all results. In general \methodname{} performs fairly well across the board. 
The combination of \methodname{} and Cutmix leads to the best performance on most of the metrics.

\begin{table}[h]
    \centering
    \input{tables/imagenet_a}
    \vspace{-0.2in}
\end{table}

\subsection{Beyond Computer Vision}

\label{subsec:speech_command}

We also found that MoEx can be beneficial in other areas such as natural language processing and speech recognition.
We use the Speech Command dataset\footnote{We attribute the Speech Command dataset to the Tensorflow team and AIY project: \href{https://ai.googleblog.com/2017/08/launching-speech-commands-dataset.html}{https://ai.googleblog.com/2017/08/launching-speech-commands-dataset.html}}~\cite{warden2018speech} which contains 65000 utterances (one second long) from thousands of people. The goal is to classify them in to 30 command words such as "Go", "Stop", etc. There are 56196, 7477, and 6835 examples for training, validation, and test. We use an open source implementation\footnote{\href{https://github.com/tugstugi/pytorch-speech-commands}{https://github.com/tugstugi/pytorch-speech-commands}} to encode each audio into a mel-spectrogram of size 1x32x32 and feeds it to 2D ConvNets as an one-channel input. We follow the default setup in the codebase training models with initial learning rate 0.01 with ADAM~\cite{kingma2014adam} for 70 epochs. The learning rate is reduce on plateau. We use the validation set for hyper-parameter selection and tune \methodname{} $p \in \{0.25, 0.5, 0.75, 1\}$ and $\lambda \in \{0.5, 0.9\}$. We test the proposed \methodname{} on three baselines models: DenseNet-BC-100, VGG-11-BN, and WRN-28-10.

\autoref{tab:speech} displays the validation and test errors.
We observe that training models with \methodname{} improve over the baselines significantly in all but one case. The only exception is DenseNet-BC-100, which has only  2\% of  the  parameters of the wide ResNet,  confirming the findings of \citet{zhang2017mixup} that on this data set data augmentation has little effect on tiny models.

\begin{table}[h]
    \centering
    \input{tables/speech_result}
    \vspace{-0.1in}
\end{table}

In addition, please see Appendix for additional natural language processing results. In contrast to prior augmentation methods, which combine two images in pixel or feature space through linear or non-linear interpolation, MoEx extracts and injects (higher order) statistics about the features.Moments are a natural first choice, but other statistics are possible (e.g.  principal components). 

%% file: tables/label_smoothing_ablation.tex
\footnotesize
\begin{tabular}{l|c|c}
\toprule
name & \methodname{} & Test Error \\ 
\midrule
Baseline & \xmark & 26.3$\pm$0.10\\
Label smoothing \citep{szegedy2016rethinking} & \xmark & 26.0$\pm$0.06 \\
Label  Interpolation only & \xmark & 26.0$\pm$0.12\\
\midrule
\methodname{} ($\lambda$ = 1, not interpolating the labels) & \cmark & 26.3$\pm$0.02 \\
\methodname{} with label smoothing & \cmark & 25.8$\pm$0.09 \\
\methodname{} ($\lambda$ = 0.9, label interpolation, proposed) & \cmark & \textbf{25.5$\pm$0.09} \\
\bottomrule
\end{tabular}
\caption{Ablation study on different design choices.}
\label{tab:label_smoothing_ablation}

%% file: tables/moex_norm_ablation.tex
\footnotesize
\begin{tabular}{l|c}
\toprule
Moments to exchange & Test Error \\ 
\midrule
\midrule
No \methodname{} & 26.3$\pm$0.10\\
\midrule
All features in a layer, i.e. LN & 25.6$\pm$0.02 \\
Feature in each channel, i.e. IN & 25.7$\pm$0.13 \\
Features in Group of channels, i.e. GN (g=4) & 25.7$\pm$0.09 \\
Features at each position, i.e. PONO & \textbf{25.5$\pm$0.09} \\
1st moment at each position & 25.9$\pm$0.06  \\
2nd moment at each position & 26.0$\pm$0.13  \\
Unnormalized 2nd moment at each position, i.e. LRN & 26.3$\pm$0.05 \\
\bottomrule
\end{tabular}
\caption{\methodname{} with different normalization methods on CIFAR-100.
}
\label{tab:moex_norm_ablation}

%% file: tables/ablation_norm_layer.tex
\small
\begin{tabular}{l|c|c|c}
    \toprule
    Model & Before Stage 1 & Before Stage 2 & Before Stage 3\\
    \midrule
    \midrule
    LN & 25.9 $\pm$ 0.08 & 25.9 $\pm$ 0.07 & \textbf{25.6 $\pm$ 0.02} \\
    IN & 26.0 $\pm$ 0.13 & \textbf{25.7 $\pm$ 0.13} & 26.2 $\pm$ 0.13 \\
    GN & \textbf{25.7 $\pm$ 0.09} & 26.1 $\pm$ 0.09 & 25.8 $\pm$ 0.13 \\
    PONO & \textbf{25.5 $\pm$ 0.09} & 26.1 $\pm$ 0.03 & 26.0 $\pm$ 0.09 \\
    \bottomrule
\end{tabular}
\caption{MoEx with different normalization methods applied to different layers in a 3-stage ResNet-110 on CIFAR-100. We bold the best layer of each normalization method.}
\label{tab:ablation_norm_layer}

%% file: tables/ablation_choices.tex
\begin{tabular}{c|c|c|c}
\toprule
Model &  $\lambda$&$p$ & Top-1 / Top-5 Error(\%)  \\  
\midrule\midrule
\multirow{8}{*}{ResNet-50}
 & 1 &0  & 23.1 / 6.6 \\
 & 0.9 &0.25 & 22.6 / 6.6 \\
 & 0.9 &0.5 &  22.4 / 6.4\\
 & 0.9 &0.75 & 22.3 / 6.3 \\
&0.3 & 1 &22.9 / 6.9  \\
& 0.5 &1 &22.2 / 6.4  \\
&0.7 & 1& \textbf{21.9} / 6.2 \\
& 0.9 & 1& \textbf{21.9} / \textbf{6.1}  \\
& 0.95 & 1& 22.5 /  6.3 \\
& 0.99 & 1& 22.6 / 6.5  \\
\bottomrule
\end{tabular}
\caption{Ablation study on ImageNet with different $\lambda$ and $p$ (exchange probability) trained for 300 epochs.}
\label{tab:ablation_choices}

%% file: tables/imagenet_a.tex
\resizebox{\linewidth}{!}{%
    \begin{tabular}{l|ccccc}
        \toprule
        Name & Acc$\uparrow$ & RMS$\downarrow$ & MAD$\downarrow$ & AURRA$\uparrow$ & Soft F1$\uparrow$ \\
        \midrule
        \midrule
        ResNet-50 (torchvision) & 0 & 62.6 & 55.8 & 0 & 60.0 \\
        Shape-ResNet & 2.3 & 57.8 & 50.7 & 1.8 & 62.1 \\
        AugMix & 3.8 & 51.1 & 43.7 & 3.3 & 66.8 \\
        Fast AutoAugment & 4.7 & 54.7 & 47.8 & 4.5 & 62.3 \\
        Cutout & 4.4 & 55.7 & 48.7 & 3.8 & 61.7 \\
        Mixup & 6.6 & 51.8 & 44.4 & 7.0 & 63.7 \\
        Cutmix & 7.3 & 45.0 & 36.5 & 7.2 & 69.3 \\
        \midrule
        ResNet-50 (300 epochs) & 4.2 & 54.0 & 46.8 & 3.9 & 63.7 \\
        \methodname{} & 5.5 & 43.2 & \textbf{34.2} & 5.7 & \textbf{72.9} \\
        Cutmix + \methodname{} & \textbf{7.9} & \textbf{42.6} & 34.3 & \textbf{8.5} & 70.5 \\
        \bottomrule
    \end{tabular}
}
\caption{The performance of ResNet-50 variants on ImageNet-A. The up-arrow represents the higher the better, and vice versa. }
\label{tab:imagenet_a}

%% file: tables/speech_result.tex
\begin{tabular}{l|r|c|c}
    \toprule
    Model & \# Param & Val Err & Test Err \\
    \midrule
    \midrule
    DenseNet-BC-100 & 0.8M & 3.16 & \textbf{3.23} \\
    +\methodname{} & 0.8M & \textbf{2.97} & 3.31 \\
    \midrule
    VGG-11-BN & 28.2M  & 3.05 & 3.38 \\
    +\methodname{} & 28.2M  & \textbf{2.76} & \textbf{3.00}  \\
    \midrule
    WRN-28-10 & 36.5M & 2.42 & 2.21 \\
    +\methodname{} & 36.5M & \textbf{2.22} & \textbf{1.98} \\
    \bottomrule
\end{tabular}
\caption{Speech classification on Speech Command. Similar to the observation of~\cite{zhang2017mixup}, regularization methods work better for models with large capacity on this dataset.}
\label{tab:speech}

%% file: sections/conclusion.tex
\section{Conclusion and Future Work}
\label{sec:conclusion}
In this paper we propose \methodname{}, a novel data augmentation algorithm for deep recognition models. Instead of disregarding the moments extracted by the (intra-instance) normalization layer, it forces the neural network to pay special attention towards them. 
We show empirically that this approach is consistently able to improve classification accuracy and robustness across many data sets, model architectures, and prediction tasks. As an augmentation method in feature space, \methodname{} is complementary to existing state-of-the-art approaches and can be readily combined with them. Because of its ease of use and extremely simple implementation 
we hope that \methodname{} will be useful to many  
practitioners in computer vision, and beyond --- in fact, anybody who trains discriminative deep networks with mini-batches.

%% file: sections/acknowledgement.tex
\subsection*{Acknowledgments}
This research is supported in part by the grants from Facebook, DARPA, the National Science Foundation (III-1618134, III-1526012, IIS1149882, IIS-1724282, and TRIPODS-1740822), the Office of Naval Research DOD (N00014-17-1-2175), Bill and Melinda Gates Foundation. We are thankful for generous support by Zillow and SAP America Inc. Facebook has no collaboration with the other sponsors of this project. In particular, we appreciate the valuable discussion with Gao Huang.

%% file: sections/appendix.tex
\section{\methodname{} PyTorch Implementation}
\label{moex_pytorch_implementation}
\autoref{alg:pytorch} shows an example code of \methodname{} in PyTorch~\citep{paszke2017automatic}.

\definecolor{backcolour}{RGB}{250,250,250}
\definecolor{keywords}{RGB}{255,0,90} 
\definecolor{comments}{RGB}{0,0,113}
\definecolor{codered}{RGB}{160,0,0}
\definecolor{codegreen}{RGB}{0,150,0}
\definecolor{codeblue}{RGB}{0,90,255}
\definecolor{codeorange}{RGB}{255,90,0}

\lstdefinestyle{mypython}{language=Python, 
    basicstyle=\ttfamily\scriptsize, 
    backgroundcolor=\color{backcolour}, 
    keywordstyle=\color{codeorange},
    commentstyle=\color{comments},
    stringstyle=\color{codered},
    showstringspaces=false,
    identifierstyle=\color{codegreen},              
    captionpos=b,
    procnamekeys={def,class}
}

\begin{lstlisting}[language=Python, label={alg:pytorch}, caption={Example code of \methodname{} in PyTorch.},style=mypython]
# x: a batch of features of shape (batch_size, 
#    channels, height, width),
# y: onehot labels of shape (batch_size, n_classes)
# norm_type: type of the normalization to use

def moex(x, y, norm_type):
    x, mean, std = normalization(x, norm_type)
    ex_index = torch.randperm(x.shape[0])
    x = x * std[ex_index] + mean[ex_index]
    y_b = y[ex_index]
    return x, y, y_b
    
# output: model output
# y: original labels
# y_b: labels of moments
# loss_func: loss function used originally
# lam: interpolation weight $\lambda$
def interpolate_loss(output, y, y_b, loss_func, lam):
    return lam * loss_func(output, y) + \
        (1. - lam) * loss_func(output, y_b)
    
def normalization(x, norm_type, epsilon=1e-5):
    # decide how to compute the moments
    if norm_type == 'pono':
        norm_dims = [1]
    elif norm_type == 'instance_norm':
        norm_dims = [2, 3]
    else: # layer norm
        norm_dims = [1, 2, 3]
    # compute the moments
    mean = x.mean(dim=norm_dims, keepdim=True)
    var = x.var(dim=norm_dims, keepdim=True)
    std = (var + epsilon).sqrt()
    # normalize the features, i.e., remove the moments
    x = (x - mean) / std
    return x, mean, std
\end{lstlisting}

\section{MoEx for NLP}
\label{sec:speech_nlp}
\subsection{Machine Translation on IWSLT 2014}
\label{subsec:iwslt}

To show the potential of \methodname{} on natural language processing (NLP) tasks, we apply \methodname{} to the state-of-the-art DynamicConv~\cite{wu2019pay} model on 4 tasks in a benchmarking dataset IWSLT 2014~\cite{cettolo2014report}: German to English, English to German, Italian to English, and English to Italian machine translation. 
IWSLT 2014 is based on the transcripts of TED talks and their translation, it contains 167K English and German sentence pairs and 175K English and Italian sentence pairs. We use fairseq library~\cite{ott2019fairseq} and follow the common setup~\cite{edunov2018classical} using 1/23 of the full training set as the validation set for hyper-parameter selection and early stopping.
All models are trained with a batch size of 12000 tokens per GPU on 4 GPUs for 20K updates to ensure convergence; however, the models usually don't improve after 10K updates. 
We use the validation set to select the best model. 
We tune the hyper-parameters of \methodname{} on the validation set of the German to English task including $p \in \{0.25, 0.5, 0.75, 1.0\}$ and $\lambda \in \{0.4, 0.5, 0.6, 0.7, 0.8, 0.9\}$ and  use \methodname{} with InstanceNorm with $p = 0.5$ and $\lambda = 0.8$ after the first encoder layer. We apply the same set of hyper-parameters to the other three language pairs. 
When computing the moments, the edge paddings are ignored.
We use two metrics to evaluate the models: BLEU~\cite{papineni2002bleu} which is a exact word-matching metric and scaled BERTScore F1
~\cite{bert-score}.
\autoref{tab:iwslt} summarizes the average scores (higher better) with standard error rates over three runs. It shows that \methodname{} consistently improves the baseline model on all four tasks by about 0.2 BLEU and 0.2\% BERT-F1. Although these improvements are not exorbitant, they are highly consistent and, as far as we know, \methodname{} is the first label-perturbing data augmentation method that improves machine translation models.

\begin{table}[t]
    \centering
    \input{tables/iwslt_result}
\end{table}

\section{More Examples of \methodname{}}

\begin{figure*}[t]
    \centering
    \includegraphics[width=\linewidth]{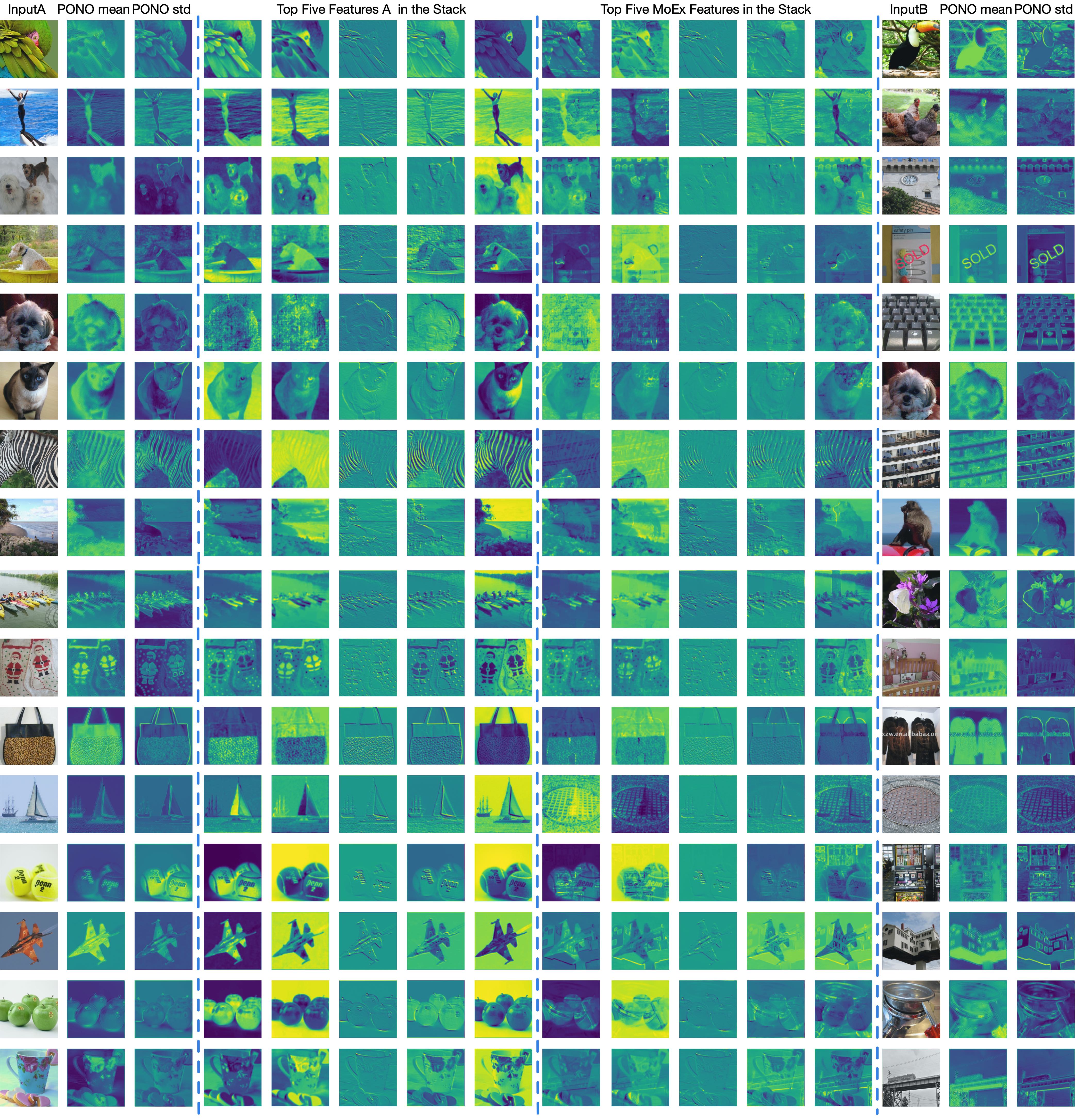}
    \caption{\methodname{} with PONO normalization. The features of image A are normalized and then infused with moments $\boldsymbol{\mu}_B$ (PONO mean), $\boldsymbol{\sigma}_B$ (PONO std) from the image B.}
    \label{fig:more_moex}
\end{figure*}

~\autoref{fig:more_moex} shows more examples of \methodname{}. We select top five features out of 64 channels to show here.

%% file: tables/iwslt_result.tex
\footnotesize
\begin{tabular}{c|l|c|c}
    \toprule
    Task & Method & BLEU $\uparrow$ & BERT-F1 (\%) $\uparrow$\\
    \midrule
    \midrule
    \multirow{4}{*}{De-En}
    & Transformer & 34.4$^\dagger$ & - \\
    & DynamicConv & 35.2$^\dagger$ & -  \\
    & DynamicConv & 35.46$\pm$0.06 & 67.28$\pm$0.02 \\
    & + \methodname{} & \textbf{35.64$\pm$0.11} & \textbf{67.44$\pm$0.09}\\
    \midrule
    \multirow{2}{*}{En-De}
    & DynamicConv & 28.96$\pm$0.05 & 63.75$\pm$0.04 \\
    & + \methodname{} & \textbf{29.18$\pm$0.10} & \textbf{63.86$\pm$0.02}\\
    \midrule
    \multirow{2}{*}{It-En}
    & DynamicConv & 33.27$\pm$0.04 & 65.51$\pm$0.02 \\
    & + \methodname{} & \textbf{33.36$\pm$0.11} & \textbf{65.65$\pm$0.07}\\
    \midrule
    \multirow{2}{*}{En-It}
    & DynamicConv & 30.47$\pm$0.06 & 64.05$\pm$0.01 \\
    & + \methodname{} & \textbf{30.64$\pm$0.06} & \textbf{64.21$\pm$0.11}\\
    \bottomrule
\end{tabular}
\caption{Machine translation with DynamicConv~\cite{wu2019pay} on IWSLT-14 German to English, English to German, Italian to English, and English to Italian tasks. The mean and standard error are based on 3 random runs.
$^\dagger$: numbers from \cite{wu2019pay}. Note: for all these scores, the higher the better.}
\label{tab:iwslt}